\documentclass[letterpaper, 10 pt, conference]{ieeeconf}
\IEEEoverridecommandlockouts
\usepackage{times}
\usepackage{dsfont}
\usepackage{multicol}
\usepackage[bookmarks=true]{hyperref}
\hypersetup{
    colorlinks=false,
    linkcolor=blue,
    filecolor=magenta,      
    urlcolor=cyan,
    pdftitle={Overleaf Example},
    pdfpagemode=FullScreen,
    }
\usepackage[utf8]{inputenc} 
\usepackage[T1]{fontenc}    
\usepackage{hyperref}       
\usepackage{url}            
\usepackage{booktabs}       
\usepackage{amsfonts,bm}       
\usepackage{amssymb}
\usepackage{nicefrac}       
\usepackage{xcolor}         
\usepackage{bbm}

\usepackage{caption}
\usepackage{subcaption}

\usepackage{bbm}
\usepackage{lipsum}
\usepackage{algorithm}
\usepackage{algpseudocode}
\usepackage{graphicx}
\usepackage{framed}
\usepackage[normalem]{ulem}
\usepackage{siunitx}

\let\labelindent\relax
\usepackage{enumitem}

\usepackage{mathtools}
\graphicspath{ {images/} }

\usepackage{multicol}
\usepackage{svg}
\usepackage{multirow}

\usepackage[normalem]{ulem}

\newcommand{\rev}[1]{\textcolor{black}{#1}}
\usepackage{xcolor}
\renewcommand\fbox{\fcolorbox{red}{white}}

\title{\LARGE \bf
Inverse Constraint Learning and Generalization \\
by Transferable Reward Decomposition
}

\author{Jaehwi Jang, Minjae Song, and Daehyung Park\textsuperscript{\textdagger}
\thanks{This work was partly supported by Institute of Information \& communications Technology Planning \& Evaluation (IITP) grant funded by the Korea government (MSIT) (No.2022-0-00311), National Research Foundation of Korea (NRF) grants funded by the Korea government (MSIT) (No.2021R1A4A3032834 and 2021R1C1C1004368), and the KAIST Convergence Research Institute Operation Program.}
\thanks{All authors are with Korea Advanced Institute of Science and Technology, Korea ({\tt\footnotesize \{wognl0402, smj0398, daehyung\}@kaist.ac.kr}). {\textsuperscript{\textdagger}}D. Park is the corresponding author.}
}
\begin{document}



\maketitle
\thispagestyle{empty} 
\pagestyle{empty}


\begin{abstract}
We present the problem of inverse constraint learning (ICL), which recovers constraints from demonstrations to autonomously reproduce constrained skills in new scenarios. However, ICL suffers from an ill-posed nature, leading to inaccurate inference of constraints from demonstrations. To figure it out, we introduce a transferable constraint learning (TCL) algorithm that jointly infers a task-oriented reward and a task-agnostic constraint, enabling the generalization of learned skills. Our method TCL additively decomposes the overall reward into a task reward and its residual as soft constraints, maximizing policy divergence between task- and constraint-oriented policies to obtain a transferable constraint. Evaluating our method and five baselines in three simulated environments, we show TCL outperforms state-of-the-art IRL and ICL algorithms, achieving up to a $72\%$ higher task-success rates with accurate decomposition compared to the next best approach in novel scenarios. Further, we demonstrate the robustness of TCL on two real-world robotic tasks.

\end{abstract}


\IEEEpeerreviewmaketitle

\section{Introduction}
Reinforcement learning (RL) has demonstrated its effectiveness in a variety of domains. However, applying these advances to real-world tasks can be challenging due to personal, task-related, and environmental restrictions in practice. Therefore, it is crucial to establish necessary constraints that impose strict limitations to ensure the validity of the task solution, independent of the optimization objective in RL. 

In this context, we aim to address the problem of inverse constraint learning (ICL), a variant of inverse reinforcement learning (IRL), that recovers constraints encoded in demonstrations to autonomously define and reuse constraints. We particularly seek transferable constraints that enable agents to reproduce a family of constrained behaviors in new scenarios, maximizing new task objectives through constrained RL (CRL)~\cite{altman1999constrained, tessler2018reward}. However, the ICL problem is ill-posed since there can be multiple possible reward-and-constraint pairs for which the reproduced behavior is optimal, leading to inaccurate inferences of underlying constraints. 

To resolve the ill-posedness, researchers have often restricted the space of constraints~\cite{armesto2017efficient, scobee2019maximum}. 
Conventional ICL methods often assume a parameterized form of constraints, such as linear (or nonlinear) equality or inequality constraints~\cite{armesto2017efficient, chou2021learning}, or a combination of constraint templates~\cite{perez2017c, park2020inferring}.
Alternatively, recent approaches expand the form of constraints by adopting high expressiveness of neural networks over discrete state-action space \cite{scobee2019maximum, mcpherson2021maximum} or continuous state-action space \cite{malik2021inverse, gaurav2023learning, liu2023benchmarking}.
However, these methods require predefining task rewards that cannot be explicitly given from demonstrations. 
Therefore, a desired method needs to recover constraints and task rewards while keeping the expressiveness of the constraint model.

\begin{figure}[t]
 \centering
  \includegraphics[width=1.0\columnwidth]{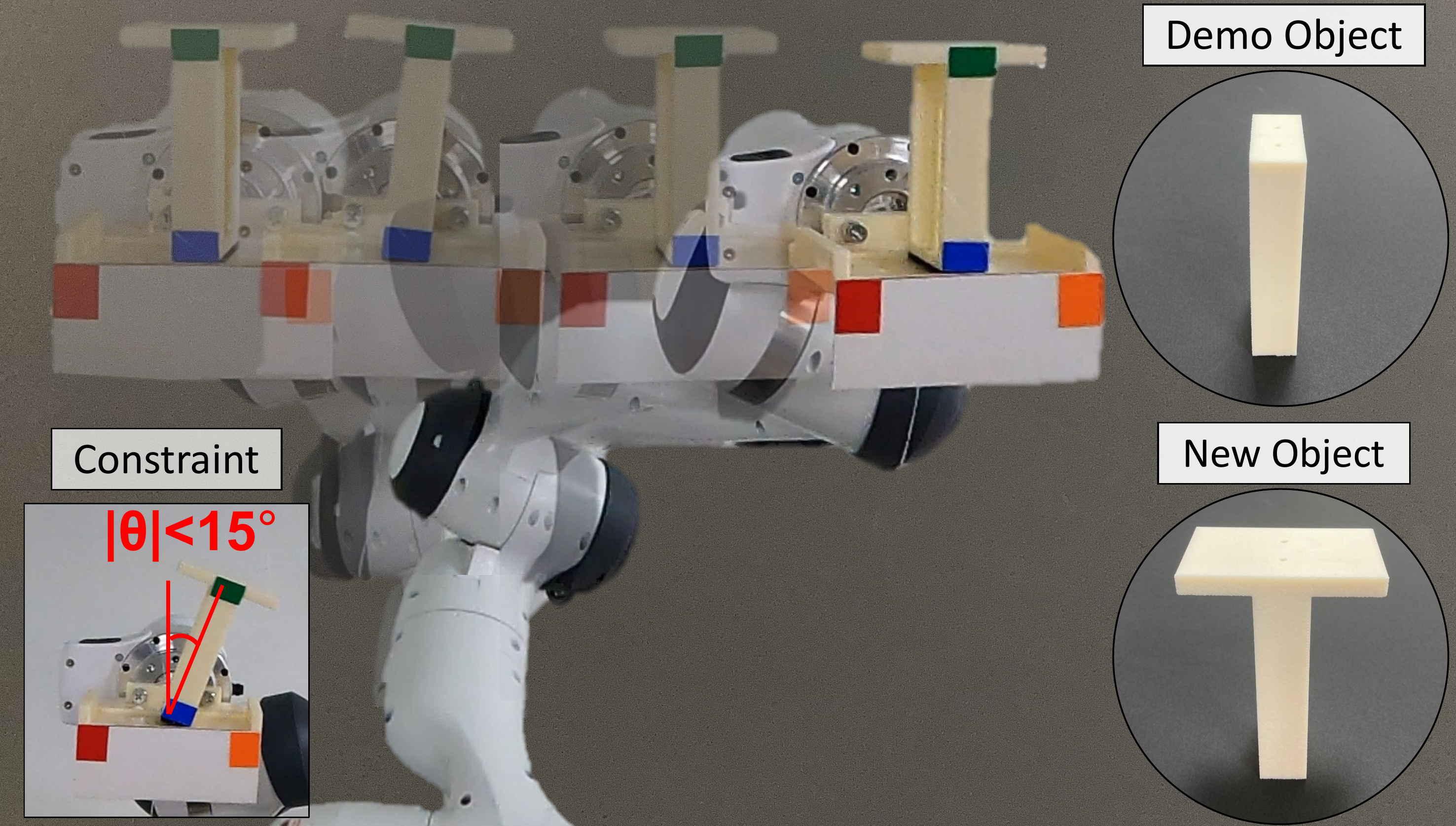}
  \caption{Illustration of a tray-carrying task, which requires transporting an object on the serving tray while preventing the object from falling, particularly constraining the angular difference from the upright state in \SI{15}{\degree}. By learning a transferable constraint from demonstrations with a \textit{demo object}, our method can generate demonstration-like constrained behaviors with novel dynamic properties of \textit{new objects}.}
  \label{fig:main}
  \vspace{-1.2em}
\end{figure}

We introduce a transferable-constraint learning (TCL) algorithm that jointly infers \textit{task} reward and \textit{residual} task-agnostic constraint pairs from demonstrations. The core idea is an expert's overall reward can be additively decomposed into a \textit{task} reward and its \textit{residual} part as a soft constraint (i.e., a negative reward with a Lagrange multiplier). 
The \textit{residual} penalizes task-oriented behavior violating underlying constraints. To infer the \textit{residual} reward without a priori known \textit{task} reward, TCL's reward decomposition (RD) method 1) finds the largest \textit{task} reward minimizing the divergence between expert and task-oriented behaviors given the space of \textit{task} rewards, and 2) selects the complement of the \textit{task} reward as the \textit{residual} reward in the overall reward. 

To expedite the RD process, we also introduce a computationally efficient approximated solution measuring the divergence between energy-based policies~\cite{haarnoja2017reinforcement}. Adopting an off-policy learning scheme, TCL efficiently infers a constraint-based transferable reward function from demonstrations. We evaluate the proposed method and five baseline methods in a total of $6000$ simulated tray-carrying, wiping, and wall-following environments. TCL outperforms ICL and IRL algorithms and shows higher task-success rates in novel scenarios. Its RD also shows superior reward decomposition, giving higher accuracy. We also demonstrate the generalization capability of transferable constraints through real tray-carrying scenarios as shown in Fig.~\ref{fig:main}. 

Our main contributions are as follows:
\begin{itemize}[leftmargin=*]
    \item We propose a novel transferable-constraint learning method that resolves the ill-posedness issue through additive reward decomposition in ICL.
    \item We introduce a more accessible ICL scheme that jointly infers reward and constraint functions without requiring a predefined \textit{task} reward.
    \item We statistically evaluate the generalization performance of learned transferable constraints through diverse simulation settings. Then, we demonstrate real-world performance using a real manipulator, Panda from Franka Emika, through tray-carrying and wall-following tasks.  
\end{itemize}

\section{Related Work}
\label{sec:related}

Constraints play a crucial role in enabling robust manipulation and allowing for more complex behaviors in real-world environments~\cite{somani2016task, perez2017c,qureshi2020neural}. Researchers have solved constrained control problems by modeling a Markov decision process (MDP) with additional constraints, resulting in a constrained Markov decision process (CMDP)~\cite{altman1999constrained}, in which Safe RL (SRL)~\cite{garcia2015comprehensive} and constrained RL~\cite{qin2021density} are representative solution frameworks.

Most SRL/CRL approaches aim to find an optimal policy maximizing the expected return while restricting the cumulative constraint cost below a threshold based on CMDP~\cite{achiam2017constrained, tessler2018reward}. The difference between approaches depends on their optimization methods, such as constrained policy optimization (CPO) that defines a trust region \cite{achiam2017constrained}, or Lagrangian relaxation of constrained optimization~\cite{chow18, tessler2018reward,huang22a,LPMHPKG2022}. Following the Lagrangian relaxation, TCL also constrains state-action costs within a margin while solving a relaxed optimization problem. 

On the other hand, ICL is an extension of IRL where the objective is, opposite to the forward SRL/CRL, recovering the constraint function from demonstrations instead of inferring the reward function~\cite{scobee2019maximum, malik2021inverse, mcpherson2021maximum, papadimitriou2021bayesian}. Conventional methods often learn the parameters of geometric constraints where a linear form of constraints is given in advance~\cite{armesto2017efficient,perez2017c}. Chou et al.~\cite{chou2021learning} introduce another model-based method of learning non-linear constraints. Recent approaches show model-free ICL that captures arbitrary Markovian feature constraints where state-action features are constrained in a set~\cite{scobee2019maximum}. Likewise, Malik et al.~\cite{malik2021inverse} show a neural network-based ICL that approximates a constraint restricting a state-action set from expert demonstrations. Papadimitriou et al. extend the aforementioned method by inferring a Bayesian posterior distribution over possible constraint sets~\cite{papadimitriou2021bayesian}. These methods require not only demonstrations but also unobservable \textit{task} (i.e., nominal) rewards. However, our method assumes only knowledge of the space of \textit{task} reward, not the reward itself.
Constrained IRL is another IRL extension that identifies rewards from demonstrations in a constrained MDP, but requires a priori known constraints \cite{fischer2021sampling, schlaginhaufen23a}.

The goal of learning constraints is to generalize expert skills across a wide range of tasks and environments~\cite{subramani2018inferring}. To achieve transferable constraints, researchers use environment-agnostic representations. Park et al.~\cite{park2020inferring} jointly recover feature constraints as well as simple \textit{task} rewards via Bayesian non-parametric IRL. Willibald and Lee then extend the work by using multimodal features~\cite{willibald2022multi} to generalize the learned constraints across geometries. Alternatively, abstract representation in semantic constraints helps encode simple or complex constraints in a unified format (e.g., $\Box\neg\textit{unsafe}$~\cite{hasanbeig2022lcrl}). Our method differs in that it recovers task-agnostic constraints from an overall reward by maximizing the divergence of policies between the \textit{task} reward and the constraints.

\section{Preliminaries}
\label{sec:background}

An MDP is a tuple, $\mathcal{M}=(\mathcal{S}, \mathcal{A}, \mathcal{T}, r, \gamma)$, where the elements represent a set of states, a set of actions, a transition function $\mathcal{T}:\mathcal{S}\times\mathcal{A} \rightarrow Pr(\mathcal{S}) $ where $Pr(\mathcal{S})$ defines a set of all probability distributions over $\mathcal{S}$, a reward function $r:\mathcal{S}\times\mathcal{A} \rightarrow \mathbb{R} $, and a discount factor $\gamma\in [0,1]$, respectively. For the MDP, a policy $\pi\in \Pi$ gives an action for each state, where $\Pi$ is a set of all stationary stochastic policies. The goal of RL is to find an optimal policy $\pi^*$ that generates a trajectory $\tau$ with its maximum cumulative discounted rewards: $\pi^*=\arg\max_{\pi}\ \mathbb{E}_{\tau \sim \pi} [ \sum_{t=1}^T \gamma^t r(s_t, a_t)]$. In this work, we use $R(\tau)$ to denote the sum of discounted rewards over the trajectory $\tau = (s_1, a_1, ..., a_{T-1}, s_T, a_T)$ with a finite horizon, $T$. 

In contrast, the goal of IRL is to infer a reward $r^*$ that best explains an expert policy $\pi_E$ for demonstrated behaviors, given an MDP, $\mathcal{M}^{-} =(\mathcal{S}, \mathcal{A}, \mathcal{T}, \gamma)$, with no reward function. 
We can solve the inference problem by adopting maximum causal entropy IRL~\cite{ziebart2010modeling}:
\begin{equation}
\begin{aligned}
r^* =  \arg&\max_{r\in\mathcal{R}} \big(\min_{\pi\in\Pi}  -H(\pi) - \mathbb{E}_{\pi}[r(s,a)] \big) + \mathbb{E}_{\pi_E}[r(s,a)],
\end{aligned}\label{eq:irl}
\end{equation}
where $H(\pi) \triangleq \mathbb{E}_\pi[-\log \pi(a \mid s)]$ is a causal-entropy term \cite{bloem2014infinite} to regularize the policy search. The rollout traces of the expert policy $\pi_E$ correspond to demonstrated trajectories $\mathcal{D}_E=(\tau_1, \tau_2, ... )$. 

\begin{figure*}[t]
 \center
  \includegraphics[width=\textwidth]{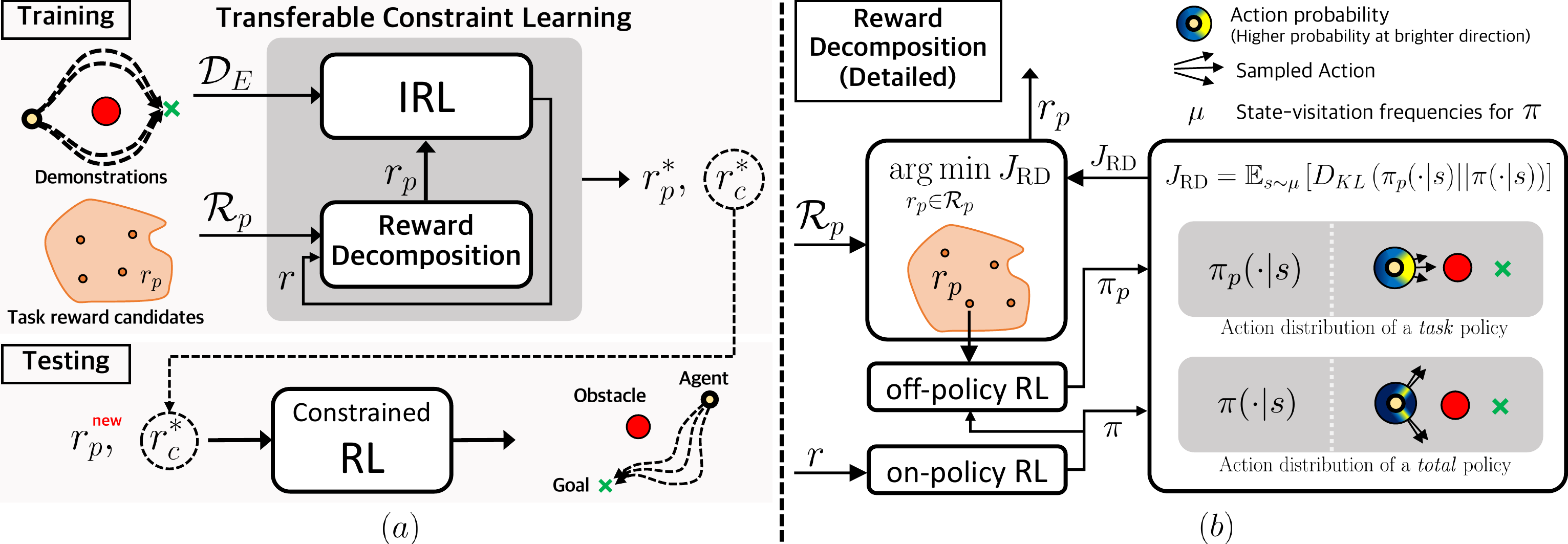}
  \caption{  The pipeline of transferable constraint learning (TCL) framework. (a) The overview of the training and testing process of TCL. In the training process, TCL takes demonstrations $\mathcal{D}_E$ and space of \textit{task} rewards $\mathcal{R}_p$, outputs a pair of \textit{task}-oriented and -agnostic rewards $(r^*_p, r^*_c)$. In the testing process, constrained reinforcement learning reproduces a demonstration-like constrained behavior using the learned \textit{task}-agnostic reward $r^*_c$ and a new \text{task} reward $r^{\text{new}}_p$. (b) The details of the reward decomposition (RD) module. In detail, an inverse reinforcement learning module infers an overall reward $r$ while the RD module decomposes $r$ into two rewards $(r_p, r_c)$. To find the task-agnostic reward $r_c$, the RD module finds $r_c$ as a complement of $r_p$ minimizing the divergence between $\pi_p$ and $\pi$.}
  \label{fig:overview}
  \vspace{-1.2em}
\end{figure*}
 
On the other hand, a CMDP~\cite{altman1999constrained} is another tuple with additional constraint, $\mathcal{M}_{c}=(\mathcal{S}, \mathcal{A}, \mathcal{T}, r, c, \gamma)$, where $c$ is a constraint-cost function $c:\mathcal{S}\times\mathcal{A} \rightarrow \mathbb{R}_{0}^{+}$, where $\mathbb{R}_{0}^{+}$ is a space of non-negative real numbers. The forward RL methods with CMDP find the optimal policy $\pi^*$, which maximizes the expected return while the expectation of cumulative constraint cost $C(\tau)=\Sigma_{t=1}^{T} c(s_t,a_t)$ over trajectory $\tau$ is less than or equal to a threshold $\xi_{\tau}\in \mathbb{R}$: $\pi^*=\arg\max_{\pi}\ \mathbb{E}_{\tau \sim \pi} [ R(\tau)]$ subject to $\mathbb{E}_{\tau\sim\pi}[C(\tau)] \leq \xi_{\tau}$ \cite{achiam2017constrained,tessler2018reward}. To relax the optimization, we can reformulate an unconstrained dual problem~\cite{tessler2018reward}  softening the constraint:
\begin{equation}
\min_{\lambda\geq0}\max_{ \pi}\ \mathbb{E}_{\tau \sim\pi} [ R(\tau) - \lambda (C(\tau)-\xi)],
\label{eq:crl}
\end{equation}
where $\lambda$ is a Lagrange multiplier. As $\lambda$ increases, the result converges to the solution of the constrained optimization problem. In this problem, $C(\tau)$ is to restrict the cumulative cost over the entire trajectory $\tau$ but does not guarantee each state is valid. Instead, we constrain an agent under $\mathcal{M}_c$ to be always in a valid state by regulating the immediate cost $c(s_t, a_t)$ at each time step $t$ with $c(s_t, a_t)\leq \xi$ at $t\in\mathbb{R}$, where $\xi\in\mathbb{R}$.

Inverse constraint learning (ICL) is then an IRL extension that seeks either a constraint or a reward-and-constraint pair from $\mathcal{M}_{c}^{-}=(\mathcal{S}, \mathcal{A}, \mathcal{T}, \gamma)$. ICL also finds an optimal policy $\pi^*$ where the expectation of constraint costs $c(s,a)$ over state-action pairs visited by the policy is less than or equal to a threshold $\xi$: $\pi^*=\arg\max_{\pi}\ \mathbb{E}_{\pi} [ r(s,a)]$ subject to $\mathbb{E}_{\pi}[c(s,a)] \leq \xi$. We will describe the detail in Sec.~\ref{sec:method}.

\section{Transferable-Constraint Learning}
\label{sec:method}

We describe the transferable-constraint learning method shown in Fig.~\ref{fig:overview}. For notational convenience, we drop the arguments of $r(s,a)$ and $c(s,a)$. We also call a reward function as `reward' or a cost function as `cost.'

\subsection{Problem Formulation}
\label{ssec:problem}

The objective of TCL is to jointly infer the \textit{task} reward $r_p \in \mathcal{R}_p$ and the constraint-related \textit{residual} reward $r_c\in\mathcal{R}$ given globally constrained demonstrations $\mathcal{D}_E$ and the space of \textit{task} rewards $\mathcal{R}_p\subseteq\mathcal{R}$. Here, $r_p$ is to produce an expert-like but unconstrained behavior, restricted in the \textit{task}-reward space $\mathcal{R}_p$. We define $r_p$ using predefined task-relevant features. $r_c$ is a negative constraint-cost function, $r_c=-c(s,a)$, which is task-agnostic and transferable to other tasks producing constrained behaviors. Note that we assume constraints are encoded in demonstrations $\mathcal{D}_E=\{\tau_1, \tau_2, ... \tau_N \}$, where each demonstration $\tau$ is a state-action trajectory and $N$ is the number of trajectories.

The core idea behind TCL is the expert's total reward $r$ can be additively decomposed into a pair of rewards, $r_p$ and $r_c$, based on $Q$-decomposition~\cite{russell2003q}. Then, we can formulate an optimization problem that not only learns an overall reward $r$ from demonstrations $\mathcal{D}_E$ via IRL but also decomposes $r$ to an optimal reward pair $(r^*_p, r^*_c)$ via RD:
\begin{align}
(r^*_p, r^*_c) =  &\mathop{\arg\max}\limits_{\substack{r_p\in\mathcal{R}_p\\\mathmakebox[10\fboxrule][r]{r_c} \in \mathcal{R}}}
 \big(\min_{\pi\in\Pi}  J_\text{IRL}(r,\pi ;\mathcal{D}_E)\big) - J_\text{RD}(r_p, \pi; \mathcal{R}_p) \nonumber \\ 
&\hspace{1.85em} \textrm{s.t.} \quad r=r_p+r_c,
\label{eq:problem} 
\end{align}
where $J_\textit{IRL}$ and $J_\text{RD}$ are objective functions (see details in Sec.~\ref{ssec:on-policy}). Note that $\pi$ is an output policy associated with the overall reward $r$. 

\subsection{Objective Functions}
\label{ssec:on-policy}
We further describe objective functions used in Eq.~(\ref{eq:problem}).

\noindent\textbf{1) $J_\text{IRL}$ objective function}:
The IRL objective function $J_{\text{IRL}}$ is to infer a total reward $r$ and its associated policy $\pi$ that can produce behaviors similar to demonstrations $\mathcal{D}_E$. Adopting the total reward $r$ on maximum casual entropy IRL~\cite{ziebart2010modeling}, we design $J_{\text{IRL}}$ as a max-min optimization objective (i.e., $\max_r\min_\pi J_{\text{IRL}}(r, \pi ; \mathcal{D}_E)$) that maximizes the reward expectation of the policy $\pi$ while minimizing the difference between the expectation and the empirical expectation around the demonstrations $\mathcal{D}_E$:
\begin{align}
J_{IRL}(r, \pi;\mathcal{D}_E) = & \mathbb{E}_{s,a \sim \mathcal{D}_E} [ r(s,a) ]
\nonumber\\
&-\mathbb{E}_{s,a\sim\pi}[ r(s,a) ] - H(\pi), 
\label{eq:j_irl}
\end{align}
where $H(\pi)=\mathbb{E}_{\pi}[-\log \pi(a|s) ]$ is the entropy regularization term to reduce the ill-posedness of multiple optimal policies in IRL. 
In this work, we use parameterized reward models, such as a linear combination of features or neural networks, that are flexible enough to represent not only the \textit{task} reward but also the \textit{residual} reward. Therefore, we can find an optimal reward $r^*$ and policy $\pi^*$ given $\mathcal{D}_E$. 

\noindent\textbf{2) $J_\text{RD}$ objective function}: The reward-decomposition function $J_\text{RD}$ is to select $r_c$ given the overall reward $r$, assuming the additive decomposition $r=r_p+r_c$. We use $r_p$ to define a \textit{task} policy $\pi_p$, where the execution of $\pi_p$ governs all the task-relevant but unconstrained actions. By finding the most extensive $r_p$ close to $r$, we can also obtain a task-agnostic \textit{residual} reward $r_c$ where its corresponding policy can be transferred to other tasks. To find such $r_p$, we design an objective function $J_\text{RD}$ minimizing the action divergence between $\pi_p$ and $\pi$: 
\begin{align}
J_\text{RD}(r_p, \pi ;\mathcal{R}_p) &= \mathbb{E}_{s \sim \mu} \big[ D_{KL} (\pi_p(\cdot|s) || \pi(\cdot | s) )\big],
\label{eq:j_rd}
\end{align} 
where $\mu$ is the state-visitation frequencies for $\pi$ and $D_{KL}(\cdot)$ represents a Kullback-Leibler (KL) divergence between two given policies. Note that $\pi_p $ optimizes the sum of $r_p$, $\pi_p = f(r_p)$. In this work, $r_p$ returns a sparse non-negative value or the highest value at the goal state to guide the agent. 

In this work, we manually choose task-relevant features, which serve as the basis vectors of the task space. In cases where task-relevant features are not readily available, we can use features suitable for sparse rewards or binary indicators.

\subsection{Generalization of Learned Constraints}
\label{ssec:generalization} 
Given the learned constraint-cost function $c=-r_c$ , we want to reproduce the demonstration-like constrained behaviors in unforeseen environments by inferring a new optimal policy $\pi^{*}_\text{new}$. 
Defining a new \textit{task} reward $r^\text{new}_p$ and a cost threshold $\xi$ of transferred constraint, $c\leq \xi$, we formulate a Lagrangian dual problem for CRL similar to Eq.~(\ref{eq:crl}):
 \begin{align}
 \min_{\lambda \geq 0} \max_{\pi} \mathbb{E}_\pi [ \underbrace{r^\text{new}_p (s,a)}_{\substack{\text{user-defined} \\ \text{task reward}}} + \underbrace{\lambda r_c(s,a) + \lambda \xi}_{\substack{\text{transferred constraint} \\ \text{-related reward}}} ],
 \end{align}
where $r^\text{new}_p \in \mathcal{R}_p$ is a user-defined \textit{task} reward for the new setup. The CRL procedure finds a high-entropy policy that maximizes the new cumulative reward. In this work, we use the inequality threshold $\xi$ that validates all samples in the expert demonstrations: $\xi = \max_{(s,a)\in \mathcal{D}_E} -r_c(s,a)$. 

\section{Approximation of TCL}
\label{sec:approx}
The optimization problem of TCL, particularly its reward decomposition, requires restricting $r_p \in \mathcal{R}_p$. In detail, Eq.~(\ref{eq:j_rd}) needs to constrain $\pi_p \in \Pi_p$ where $\Pi_p$ is a space of policies from $r_p \in \mathcal{R}_p$.
However, the condition $\pi_p \in \Pi_p$ requires non-trivial computation to obtain action-value functions $Q_p \in \mathcal{Q}_p$, where $\mathcal{Q}_p$ is a space of action-value functions given $\mathcal{R}_p$.

To reduce the computational complexity, we simplify the reward decomposition by approximating the KL divergence and its condition. Adopting energy-based policies~\cite{haarnoja2017reinforcement} that $\pi_p (\cdot|s) \propto \exp(Q_p(s, \cdot))$, we formulate a new minimization problem for the reward decomposition as follows:
\begin{align}
\min_{Q_p \in \mathcal{Q}_p} D_{KL} \Big( \frac{\exp  Q_p(s, \cdot)}{Z_p}|| \frac{\exp  Q(s, \cdot) }{Z}\Big),
\label{eq:approx}
\end{align}
where $Z_p$ and $Z$ are normalization factors, $Z_p=\Sigma_a \exp Q_p(s,a)$ and $Z=\Sigma_a \exp Q(s,a)$, respectively. 
In this work, we train the state-action value functions, $Q$ and $Q_p$ given rollouts from $\pi$, following the Bellman equations:
\begin{align}
Q(s,a) &\coloneqq r(s,a) + \gamma V(s')  \\
Q_p(s,a) &\coloneqq r_p(s,a) + \gamma V_p(s'),
\end{align}
where $V(s)$ and $V_p(s)$ are state-value functions given rewards $r$ and $r_p$, respectively. Note that we compute $V_p(s) = \mathbb{E}_{a\sim \pi(\cdot, s)} [Q_p(s,a)]$ assuming the total policy $\pi$. 

Finally, we formulate a new approximated $J_\text{RD}$ function by projecting the condition $Q_p\in\mathcal{Q}_p$ in Eq.~(\ref{eq:approx}) onto the space of rewards to reduce computational cost and also to use the existing condition $r_p\in\mathcal{R}_p$:
\begin{equation}
\begin{aligned}
J_\text{RD}(r_p, \pi;&\mathcal{R}_p)= 
D_{KL} \Big( \frac{\exp  Q_p(s, \cdot)}{Z_p}|| \frac{\exp  Q(s, \cdot) }{Z}\Big) 
 \\ 
&+ \alpha \cdot || Q_p (s,a) - \big(r_p(s,a) + \gamma V_p(s')\big) ||^2, 
\end{aligned}
\end{equation}
where $\alpha \in \mathbb{R}^{+}$ is an adjustable constant, which is $1$ in this work.
\section{Experimental Setup}
\label{sec:eval_setup}

\subsection{Quantitative Evaluation through Simulation}
\label{ssec:Quantitative}
We assess the learning and reproduction capabilities of TCL and baseline methods. Fig.~\ref{fig:exp-box2d-comparison} presents the simulated environments: \textit{tray-carrying}, \textit{wall-following}, and \textit{wiping} created using a 2-dimensional physics simulator engine, called Box2D~\cite{catto2011box2d}. For each environment, we collected expert demonstrations $\mathcal{D}_E$ and trained a rollout policy using reward-constrained policy optimization (RCPO) \cite{tessler2018reward} with manually designed constraints. Subsequently, we evaluated each method by training with $\mathcal{D}_E$ and predefined $\mathcal{R}_p$. We describe evaluation setups below:

\begin{figure*}[t]
    \centering
    \includegraphics[width=0.98\textwidth]{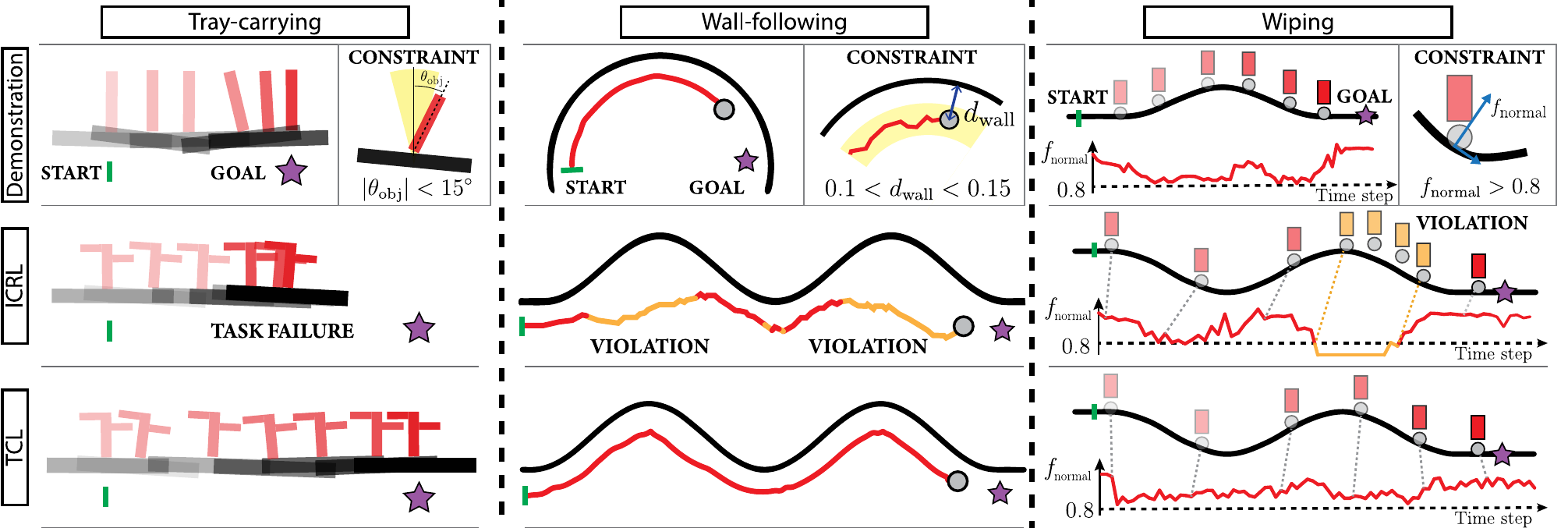}
    \caption{Comparison of constraint-transfer performance in simulated environments. In each task (column), ICRL (the second row) and TCL (the third row) first learn constraints from demonstrations (the first row). We then make a CRL-based policy trained with the learned constraints to reproduce demonstration-like behaviors in novel environments. The green and purple shapes represent start and goal locations (or configurations), respectively. The red and orange trajectories (or boxes) represent a sequence of constrained and constraint-violated behaviors, respectively.  
    } 
    \label{fig:exp-box2d-comparison}
    \vspace{-1.0em}
\end{figure*}

\begin{figure}[t]
 \center
  \includegraphics[width=1.0\columnwidth]{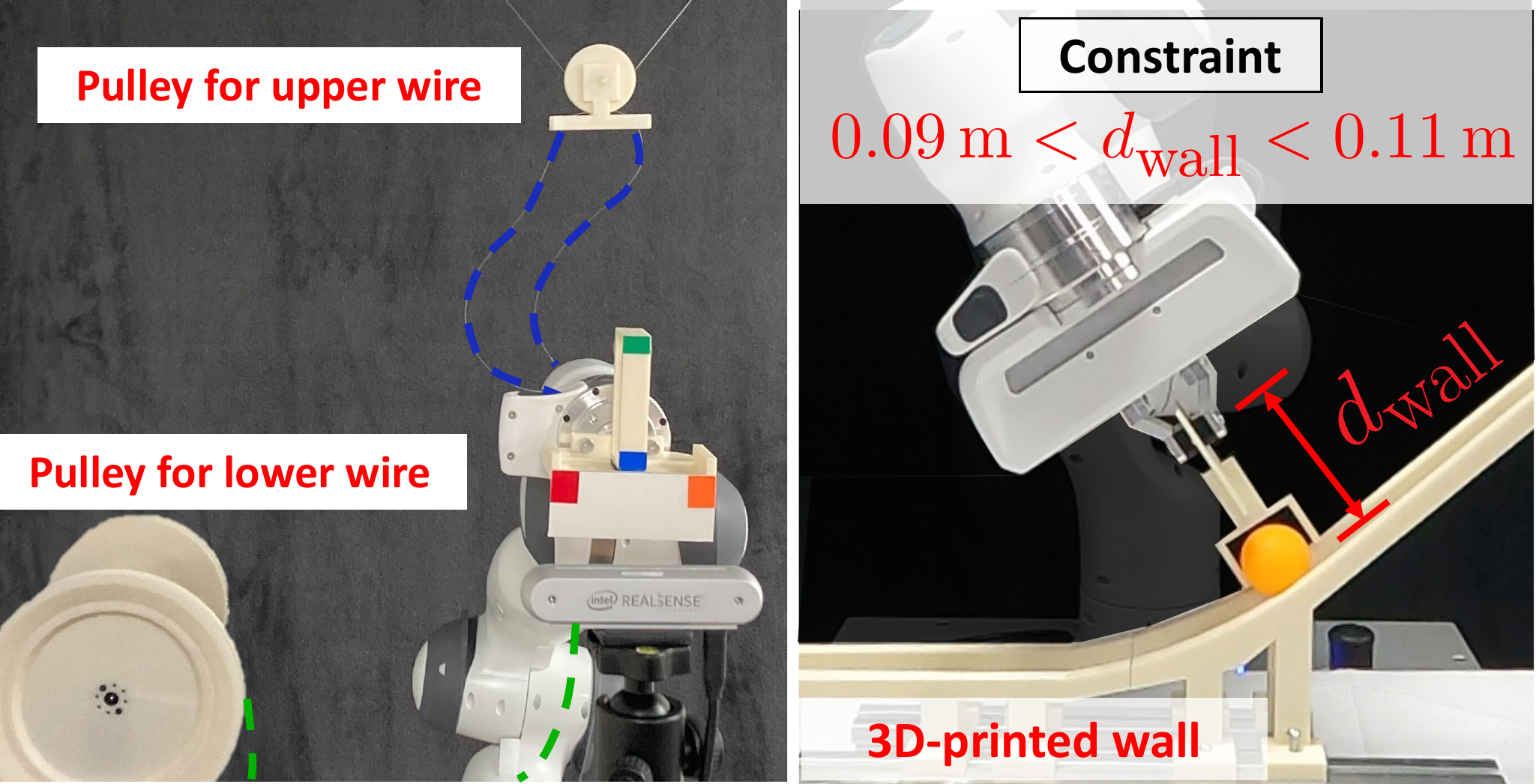}
  \caption{\rev{Two real-world environments. (\textbf{Left})} A capture of the \rev{\textit{tray-carrying} environment} with a motorized reset mechanism. A Panda arm with an attached tray moves the \rev{bar-shaped} object following a policy\rev{, where an RGB camera with an} image-based pose detector \rev{provides} the tray and an object \rev{poses in real-time}. When the object falls over a certain angle, the reset mechanism pulls up and down wires to restore the object on the tray. \rev{(\textbf{Right}) A capture of the \textit{wall-following} environment with a 3D-printed wall. The Panda arm moves along the curved wall without dropping the ball and hitting the wall.}}
  \vspace{-1.2em}
  \label{fig:reset}
\end{figure}

\noindent\textbf{1) \textit{Tray-carrying} environment}:
Fig.~\ref{fig:exp-box2d-comparison} (Left) shows the environment, which requires a tray agent (black) to move an object (red) from a start position (green) to a goal position (purple) without tilting the object more than \SI{15}{\degree}. 
Using a bar object, we trained a rollout policy network with a size of $(64,64)$ using multi-layer perceptrons (MLP) to define a CMDP with the following specifications:
\begin{itemize}[leftmargin=*]
\item $\mathcal{S}= \{s | s=[\mathbf{x}_\text{tray}, \mathbf{x}_\text{obj}, x_\text{goal}] \in \mathbb{R}^{15}$\} where $\mathbf{x}_\text{tray}$ and $\mathbf{x}_\text{obj}$ are the tray and object states, respectively. Each state includes horizontal position and velocity ($[x_\text{tray}, \dot{x}_\text{tray}]\in\mathbb{R}^2$), an angular velocity ($\dot{\theta}_\text{tray}\in\mathbb{R}$), and inclination angle encodings ($[\sin(\theta_\text{tray}),\cos(\theta_\text{tray}),\sin(2\theta_\text{tray}),\cos(2\theta_\text{tray})]\in\mathbb{R}^4$). $x_\text{goal}\in[-0.6, 0.6]$  is the desired tray position.
\item $\mathcal{A}=\{a | a=[f_x, \tau_\theta] \in \mathbb{R}^{2}\}$ where $f_x \in[-0.5, 0.5]$ and $\tau_\theta \in [-0.05, 0.05]$ are the horizontal force and tilting torque, respectively.
\item $r_{E} = \mathds{1}^{d \leq 0.01}-0.1\cdot d\cdot \mathds{1}^{d > 0.01}$ where $d=||x_\text{tray} - x_\text{goal}||$
\item $c_{E} = \mathds{1}^{|\theta_\text{obj}|>\SI{15}{\degree}}$
\end{itemize}
where $r_{E}$ and $c_E$ are the expert reward and constraint, respectively. We collected $32$ demonstrations ($T=100$) for training.

For testing, we generated $6000$ environments with three types of objects, where $3000$ used demonstration-like start-and-goal positions, while the remaining $3000$ had uniform-randomly sampled positions. Note that we defined $\mathcal{R}_p$ based on two features: $x_\text{tray}$ and $x_\text{goal}$.

\noindent\textbf{2) \textit{Wall-following} environment}:
Fig.~\ref{fig:exp-box2d-comparison} (Middle) shows the environment, which requires a point agent (gray) to reach a goal (green) while maintaining a certain distance from a wall (black). The specifications are as follows:
\begin{itemize}[leftmargin=*]
\item $\mathcal{S} = \{ s | s = [x_\text{rel}, y_\text{rel}, d_\text{wall}] \in \mathbb{R}^3 \}$ where $[x_\text{rel}, y_\text{rel}]$ is the agent's displacement from a goal and $d_\text{wall}$ is the agent's displacement from the wall.
\item \rev{$\mathcal{A} = \{a | a = [\Delta x, \Delta y]\in \mathbb{R}^2\}$ where $\Delta x$ and $\Delta y$} are horizontal and vertical \rev{displacements} ($\in [-1, 1]$), respectively.
\item $r_E = \mathds{1}^{d \leq 0.01}-0.1\cdot d\cdot \mathds{1}^{d > 0.01}$ where $d=|| (x_\text{rel}, y_\text{rel})||_{2}$
\item$c_E = \mathds{1}^{ d_\text{wall} <0.1  	\lor d_\text{wall} > 0.15}$
\end{itemize}
We sampled $32$ demonstrations. For testing, we randomly generated $100$ sinusoidal curve environments using the function:
$y=A \sin^2(\pi x) + B \cos(\frac{\pi}{2} x) $ where A and B are randomly selected constants ($\in [0.1, 0.3]$). Note that we defined $\mathcal{R}_p$ with a feature $d$ only.

\noindent\textbf{3) \textit{Wiping}}:
Fig.~\ref{fig:exp-box2d-comparison} (Right) shows the environment, which requires a bar agent (gray) to reach a goal (purple) while maintaining a certain contact force from the curved ground (black). The specifications are as follows:
\begin{itemize}[leftmargin=*]
\item $\mathcal{S} = \{ s | s = [x_\text{rel}, y_\text{rel}, \dot{x}_\text{rel}, \dot{y}_\text{rel},
f_\text{normal}] \in \mathbb{R}^5 \}$ where $[x_\text{rel}, y_\text{rel}]$ and $[\dot{x}_\text{rel}, \dot{y}_\text{rel}]$ are the agent's displacement and velocity from the goal, respectively. $f_\text{normal}$ is the normal contact force from the curve.
\item $\mathcal{A} = \{a | a = [f_x, f_y]\in \mathbb{R}^2 \}$, where \rev{$f_x$ and $f_y$  are horizontal and vertical forces ($\in [-1, 1]$)}.
\item $r_E = \mathds{1}^{d \leq 0.01}-0.1\cdot d\cdot \mathds{1}^{d > 0.01}$ where $d=||x_\text{rel}||$
\item $c_E = \mathds{1}^{f_\text{normal} < 0.8}$
\end{itemize}
We sampled $32$ demonstrations. For testing, we randomly generated $100$ sinusoidal curves using the function:
$y=A \sin(\pi x)$ where $A\in [-0.3, 0.3]$. Note that we defined $\mathcal{R}_p$ with a feature $d$ only.

Through evaluations, we measured the constraint-violation rate as ${\sum_{i=1}^N \sum_{(s,a)\in \tau_i} c_E(s)} / \sum_{i=1}^N \|\tau_i\|$, where $\tau_i$ represents a test trajectory and $N$ is the number of trajectories. We also measure the task-success rate, classifying an episode as a success when the agent reaches the goal without any violation. For computation, we used Intel i9-10900K CPU processor and an NVIDIA
RTX 2060 Super GPU.

\subsection{Baseline Methods}
For evaluation, we employed \rev{five} baseline methods:
\begin{itemize}[leftmargin=*]
\item GAIL-Constraint (GC): An imitation learning method, GAIL~\cite{ho2016generative}, associated with constraints~\cite{malik2021inverse}. GC finds $r_c$ given a user-defined \textit{task} reward $r_p$. 
\item AIRL-Constraint (AC): A variant of adversarial IRL method \cite{fu2017learning} associated with constraints. AC infers $r$ that satisfies all constraints assuming a known \textit{task} reward. 
\item \rev{ICL-Soft\footnote{\rev{We changed the original acronym to ICL-Soft to distinguish it from inverse constraint learning (ICL).}}: An ICL method that finds cumulative soft constraints assuming a known reward~\cite{gaurav2023learning}}.
\item Feature-Constraint (FC): A distribution-based constraint learner used in CBN-IRL~\cite{park2020inferring}. FC uses a conjunction of feature-wise constraints that restrict current features within the observed range in demonstrations.
\item ICRL: An ICL method that finds a neural-network based $r_c$ from demonstrations and combines it with a predefined \textit{task} reward $r_p$ to obtain $r$~\cite{malik2021inverse}.
\end{itemize}

\subsection{Qualitative \rev{Evaluations} with a Real Robot}
We \rev{demonstrate} our proposed method in \rev{two} real-world \rev{manipulation environments using a $7$ degree-of-freedom robotic arm, Panda from Franka Emika, as shown in Fig.~\ref{fig:reset}.}

\rev{\noindent\textbf{1) \textit{tray-carrying}}: This is a real-world} \textit{tray-carrying} environment using a robotic arm-actuated serving tray, a 3D printed object, and an RGB camera. \rev{Placing a bar-shaped object on the tray, we train a rollout policy via RCPO and then collect 16 demonstrations ($T=30$) with randomly selected start-and-goal locations. After obtaining constraints using TCL with demonstrations in the environment, the robot agent learns and reproduces a demonstration-like behavior in novel random environments with an unforeseen T-shaped object.}

\rev{To show the generalization capability, we use a bar-shaped object for training and a T-shaped object for testing. To detect tray and object poses, we use a 2D-pose estimator with a \SI{60}{\hertz} blob-detection module from OpenCV2~\cite{bradski2000opencv}. In operation,} our RL-based control system applies horizontal force $f_x$ and tilt torque $\tau_\theta$ to the tray using a \SI{1}{\kilo\hertz} low-level torque controller and a \SI{10}{\hertz} high-level operational-space force/torque controller with the Jacobian transpose method. To expedite the learning process, we use a motorized reset mechanism that automatically restores the target object within \SI{17}{\second} in reset conditions such as failure or completion. We particularly reset when the object is tilted more than \SI{90}{\degree} or placed outside the workspace.

\rev{\noindent\textbf{2) \textit{wall-following}}: This is a real-world \textit{wall-following} environment in which a robot-attached tool moves along the curved wall without dropping the ball and hitting the wall. For training, we collect expert demonstrations maintaining the end-effector displacement from a wall, $c_E = \mathds{1}^{ d_\text{wall} < \SI{0.11}{\m}  	\lor d_\text{wall} > \SI{0.09}{\m}}$, via a Box2D simulator. Then, inferring constraints from demonstrations, we enable TCL to generate a CRL-based policy that reproduces the constrained behavior when given a new real-world wall. To expedite the learning process, we use a simulated \textit{wall-following} environment. We then transfer the learned policy to the Panda arm running in the new environment.}

For these evaluations, we used an Intel i9 CPU and a NVIDIA RTX 3070 GPU.

\begin{figure}[t]
    \centering
    \includegraphics[width=1.0\columnwidth]{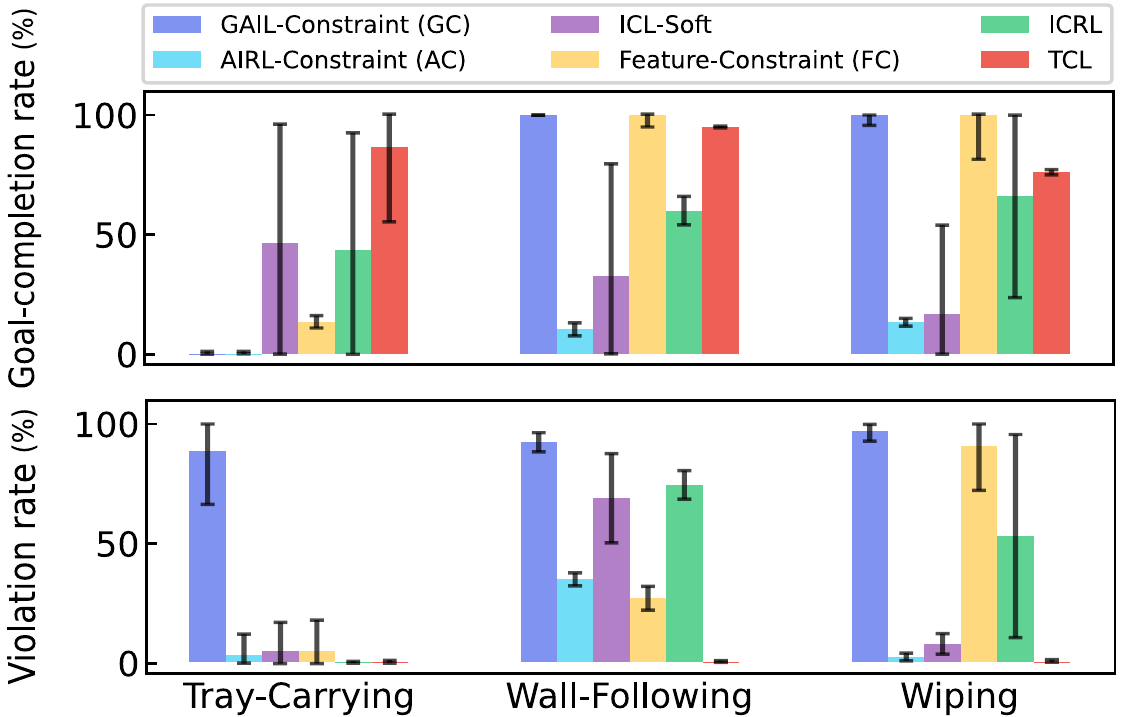}
    \caption{
    Comparison of goal-completion and constraint-violation rates between TCL and baselines in simulated environments. Note that we determine the goal-completion rate regardless of any constraint violations.
    }
    \label{fig:exp-box2d-comparison-violation}
    \vspace{-1.5em}
\end{figure}

\begin{figure*}[t]
    \centering
    \includegraphics[width=\textwidth]{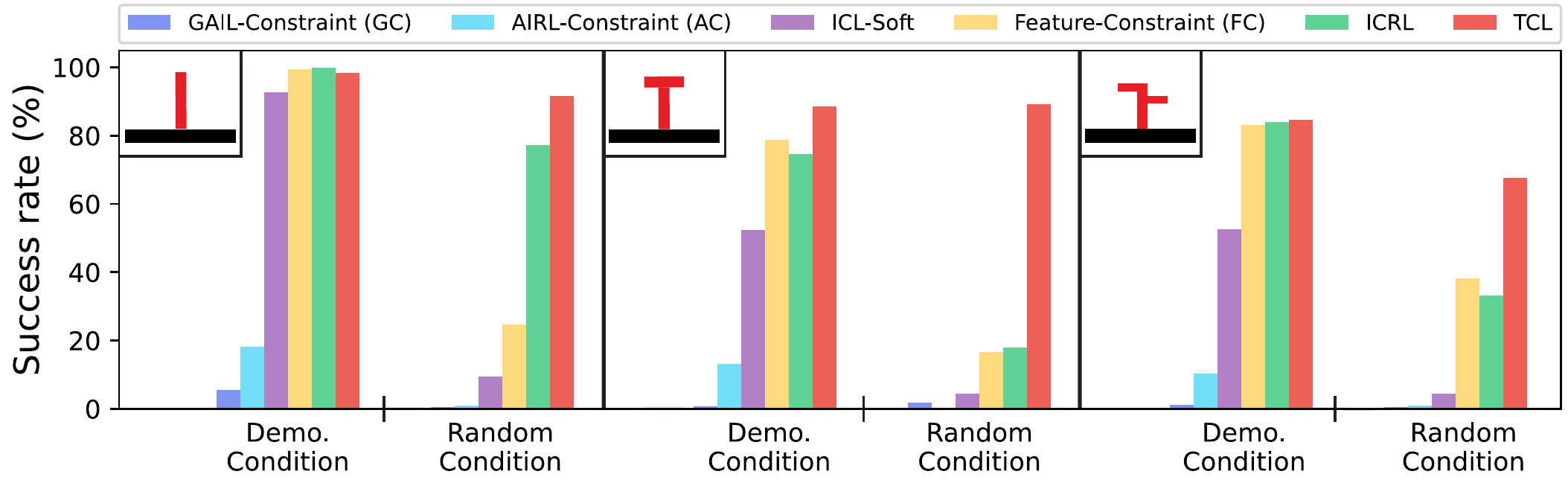}
    \caption{Comparison of task-success rates in simulated \textit{tray-carrying} environments, where we collected $32$ demonstrations with a bar-shaped object (\textbf{Left}) and reproduced demonstration-like constrained behaviors training our proposed TCL and \rev{five} baseline methods. The \textit{demonstration condition} represents the use of demonstration-like start-and goal-locations with small variations. The \textit{random condition} represents the use of uniform-randomly sampled start-and-goal locations. 
    }
    \label{fig:exp-box2d-eval}
    \vspace{-1.2em}
\end{figure*}

\section{Evaluation Results}

\begin{figure}[t]
\centering
\includegraphics[width=1.0\columnwidth]{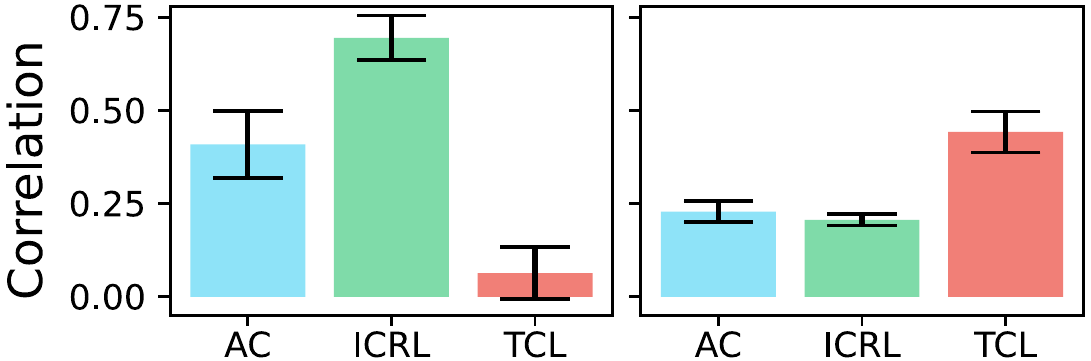}
\caption{Comparison of reward-decomposition performance over $100$ 2D reaching environments. (\textbf{Left}) The correlations between the ground-truth task-oriented reward $r_p$ and the recovered constraint-relevant reward $r_c$, and 
(\textbf{Right}) The correlations between the ground truth and the inferred constraint-relevant rewards. Note that we use a negative cost as a constraint-relevant reward. The video description can
be found here: \href{https://youtu.be/SD9HV1z5owk}{https://youtu.be/SD9HV1z5owk}}
\label{fig:corr}
\vspace{-1.2em}
\end{figure}

We analyze the constraint learning-and-transfer performance of our proposed method, TCL, and baselines in simulated environments. The qualitative study in Fig.~\ref{fig:exp-box2d-comparison} demonstrates that TCL outperforms a representative baseline, ICRL, in transferring constraints to novel environments. TCL successfully learns constraints from demonstrations and reproduces demonstration-like constrained behaviors in novel environments, as shown in the first and third rows of Fig.~\ref{fig:exp-box2d-comparison}. In the \textit{wall-following} task that requires accurate decomposition of \textit{task} and constraint-relevant \textit{residual} rewards, TCL enables the agent to reach the goal without violations, tightly following the wall against the goal-directed behavior. On the other hand, ICRL either fails to reach the goal (i.e., \textit{tray-carrying}) or violates constraints (i.e., \textit{wall-following} and \textit{wiping}) more frequently than TCL, since ICRL cannot distinguish between task rewards and task-agnostic constraints that minimally affect a new task objective in a novel environment. 

Our quantitative study in Fig.~\ref{fig:exp-box2d-comparison-violation} further demonstrates the superior generalization performance of TCL across diverse simulations. TCL consistently exhibits the lowest constraint-violation rate in all the novel test environments, while achieving high {goal-completion} rates with different task configurations and objects. In contrast, the other baselines show either low {goal-completion} rates or high violation rates. Note that the {goal-completion} rates of GC and FC in \textit{wall-following} and \textit{wiping} environments may appear high, but most of the results are not valid due to the high rate of constraint violations. Additionally, TCL showcases the capabilities of ICL by learning not only angle and distance constraints but also force constraints defined in feature space.

We conducted a more detailed statistical evaluation to measure task-success rates in randomly generated \textit{tray-carrying} environments (see Fig.~\ref{fig:exp-box2d-eval}) The results demonstrate that TCL consistently outperforms the other baselines, achieving the highest success rate (up to $89.1\%$) with a maximum $72\%$ difference from the next best approach, ICRL, when confronted two novel objects. This indicates that TCL discovers task-agnostic constraints that can be transferred to novel environments regardless of initial configurations and object dynamics. In contrast, ICRL, ICL-Soft, and FC exhibit similar performance when given demonstration-like start and goal configurations. However, when faced with new objects, the success rates of ICRL, ICL-Soft, and FC drastically decrease, suggesting that their extracted constraints still contain task-specific components that may conflict with reproducing properly constrained behaviors in a new setup. Furthermore, GC and AC consistently show the lowest performance since these methods excessively constrain state-action pairs~\cite{malik2021inverse}. AC shows similar or higher success rates than GC since AIRL can directly infer a reward. For this study, we determine a task to be successful when both the tray and the object are on the goal while the object remains upright.

We also investigated the decomposition performance of our method and the baselines by measuring the sample correlation coefficient between obtained rewards in 2D reaching environments, as shown in Fig.~\ref{fig:overview}. To compute the correlation, we first inferred reward and constraint functions per algorithm from $100$ expert demonstrations. We then sample $1000$ constraint-relevant rewards at each of the $100$ randomly generated environments. Given two sets of reward samples, we calculated the sample correlation coefficient. Fig.~\ref{fig:corr} \rev{(}\textbf{Left}\rev{)} shows TCL resulted in the lowest correlation between ground-truth task-oriented and inferred task-agnostic rewards, where the near-zero correlation bar indicates two decomposed rewards are independent. In contrast, the other baselines show exceedingly high correlations; their constraint-based rewards still contain task-dependent and non-transferable components. We also measured the correlation between the ground-truth constraint-based reward and the inferred constraint-based reward. Fig.~\ref{fig:corr} \rev{(}\textbf{Right}\rev{)} shows TCL precisely inferred constraints from demonstrations with a higher correlation than the others.

\begin{figure*}[t]
 \center
  \includegraphics[width=0.98\textwidth]{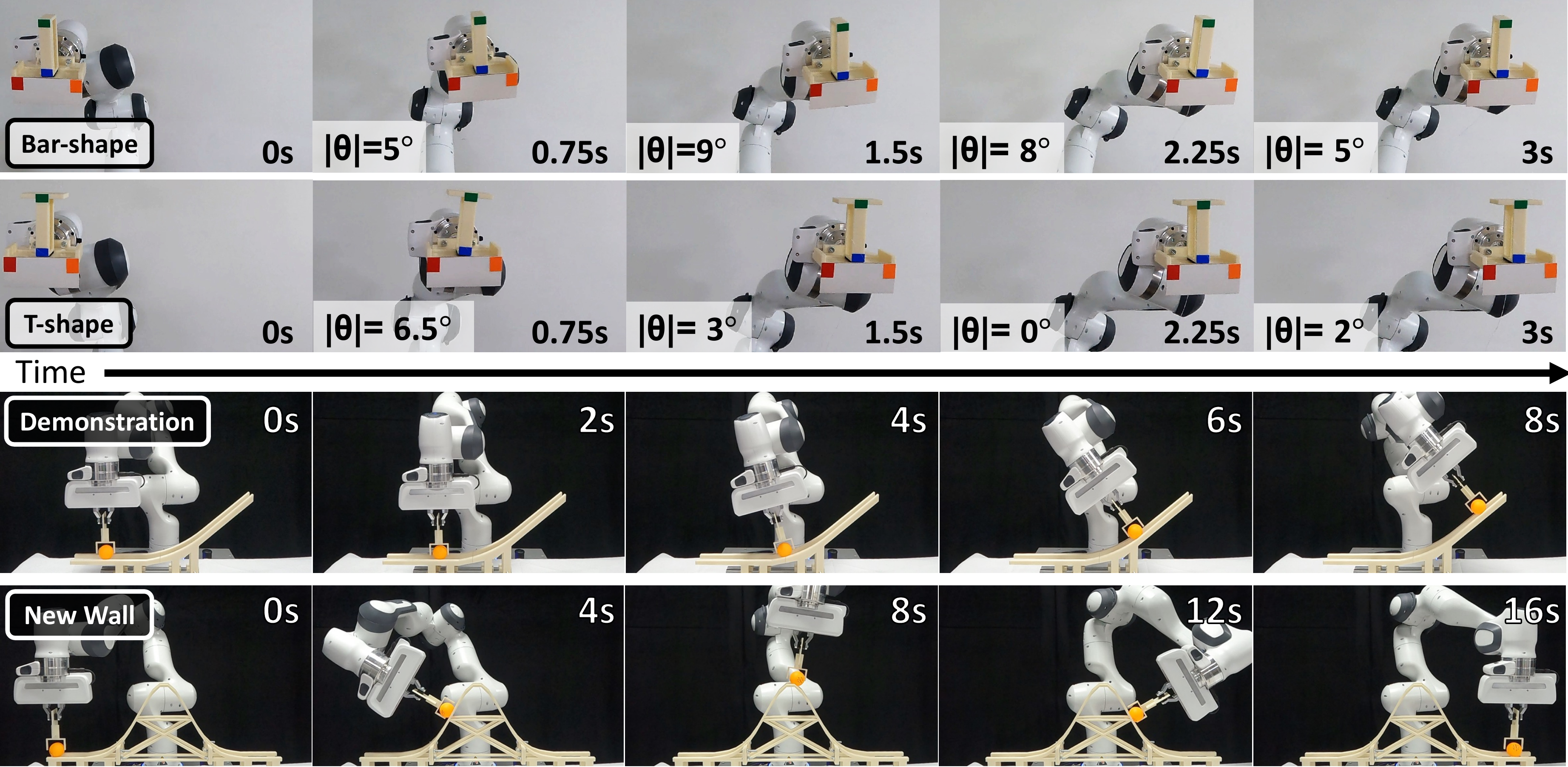}
  \caption{Demonstration of TCL's constraint-transfer capability in the  \textit{tray-carrying} \rev{and \textit{wall-following}} environments. \rev{(\textbf{Top}) In \textit{tray-carrying} environment,} TCL learns a constraint from demonstrations in the bar-shaped environment. Transferring the learned constraint, TCL then enables the robot to reproduce a demonstration-like \rev{safe carrying} behavior in the environments with an unforeseen T-shaped object. (\textbf{Bottom}) \rev{In \textit{wall-following} environment, TCL learns the transferable constraint of maintaining the distance from the wall and successfully transfers its precise collision-free tracking behavior to the new wall environment.} 
  } 
  \label{fig:sequence}
    \vspace{-1.5em}
\end{figure*}

Finally, we demonstrated the generalization performance of TCL in real-world \textit{tray-carrying} \rev{and \textit{wall-following}} environments. \rev{Fig.~\ref{fig:sequence} (Top)} shows the Panda arm with TCL successfully transported both known bar-shaped and unforeseen T-shape objects without large inclinations at every time step. \rev{Fig.~\ref{fig:sequence} (Bottom) shows TCL successfully delivered the ball moving it along the curved walls while maintaining the precise displacement constraints.} These demonstrations show TCL can robustly learn and generalize task-agnostic constraints in the real world. 
In this experiment, to obtain expert demonstrations satisfying the predefined constraint, we optimized an RCPO-based policy for 6 hours with a $30000$ step length and trained each algorithm with the real robot for $2$ hours. With the predicted constraint, our method could robustly and successfully reproduce the demonstration-like safe behavior.

\section{Conclusion}

We introduced a transferable constraint learning (TCL) algorithm that jointly infers a task-oriented reward and a task-agnostic constraint reward, as a soft constraint, from demonstrations. To accurately infer the task-agnostic reward resolving the ill-posedness of ICL, our method decomposes an overall reward to a \textit{task} and its \textit{residual} rewards following the idea of additive reward decomposition. The statistical evaluation shows our method outperforms baseline methods in terms of the task-success rate, the constraint-violation rate, the accuracy of decomposition, and the transferability to novel environments. We also demonstrate the proposed TCL can be deployed to real-world manipulation scenarios.  


\bibliographystyle{IEEEtran}
\bibliography{references}

\end{document}